\documentclass{article}

\usepackage[preprint]{Styles/neurips_2025}

\usepackage[utf8]{inputenc} 
\usepackage[T1]{fontenc}    
\usepackage{hyperref}       
\usepackage{url}            
\usepackage{booktabs}       
\usepackage{amsfonts}       
\usepackage{nicefrac}       
\usepackage{microtype}      
\usepackage{xcolor}         
\usepackage{xspace}
\usepackage{graphicx}
\usepackage{subcaption}
\usepackage{Styles/colorblock}
\usepackage{Styles/codeblock}

\newcommand{\methodname}{Prefix Grouper\xspace}

\title{\methodname: Efficient GRPO Training through Shared-Prefix Forward}

\author{%
  Zikang Liu\textsuperscript{1,2}\thanks{Equal contribution.}, Tongtian Yue\textsuperscript{1,2}\footnotemark[1], Yepeng Tang\textsuperscript{3}\footnotemark[1], \textbf{Longteng Guo\textsuperscript{1}}, \\
  \textbf{Junxian Cai\textsuperscript{4}}, \textbf{Qingbin Liu\textsuperscript{4}}, \textbf{Xi Chen\textsuperscript{4}}, \textbf{Jing Liu\textsuperscript{1,2}\thanks{Corresponding Author.}} \\
  \textsuperscript{1}Institute of Automation, Chinese Academy of Sciences \\
  \textsuperscript{2}School of Artificial Intelligence, University of Chinese Academy of Sciences \\
  \textsuperscript{3}School of Computer Science and Technology, Beijing Jiaotong University \\
  \textsuperscript{4}Basic Algorithm Center, Tencent \\
  \tt \{liuzikang2023,yuetongtian2022\}@ia.ac.cn, yepengtang@bjtu.edu.cn \\
  \tt \{jasoncjxcai,qingbinliu,jasonxchen\}@tencent.com \\
  \tt \{longteng.guo,jliu\}@nlpr.ia.ac.cn
}

\begin{document}

\maketitle

\begin{abstract}

Group Relative Policy Optimization (GRPO) enhances policy learning by computing gradients from relative comparisons among candidate outputs that share a common input prefix. 
Despite its effectiveness, GRPO introduces substantial computational overhead when processing long shared prefixes, which must be redundantly encoded for each group member. 
This inefficiency becomes a major scalability bottleneck in long-context learning scenarios.
We propose \textbf{\methodname}, an efficient GRPO training algorithm that eliminates redundant prefix computation via a \textit{Shared-Prefix Forward} strategy. In particular, by restructuring self-attention into two parts, our method enables the shared prefix to be encoded only once, while preserving full differentiability and compatibility with end-to-end training.
We provide both theoretical and empirical evidence that \methodname is training-equivalent to standard GRPO: it yields identical forward outputs and backward gradients, ensuring that the optimization dynamics and final policy performance remain unchanged. Empirically, our experiments confirm that \methodname achieves consistent results while significantly reducing the computational cost of training, particularly in long-prefix scenarios.
The proposed method is fully plug-and-play: it is compatible with existing GRPO-based architectures and can be seamlessly integrated into current training pipelines as a drop-in replacement, requiring no structural modifications and only minimal changes to input construction and attention computation.
\methodname enables the use of larger group sizes under the same computational budget, thereby improving the scalability of GRPO to more complex tasks and larger models.
Code is now available at \url{https://github.com/johncaged/PrefixGrouper}.
\end{abstract}

\section{Introduction}

Group Relative Policy Optimization (GRPO)~\cite{shao2024deepseekmath} has emerged as an effective framework for optimizing large language models in reinforcement learning. 
It avoids explicit value function estimation by comparing multiple candidate outputs generated from the same input and updating the policy based on their relative rankings.
This group-based formulation reduces gradient variance and improves training stability. GRPO has been widely applied to large-scale reasoning tasks, such as instruction following and chain-of-thought learning~\cite{yang2024qwen2, jin2025search, shen2025vlm, huang2025vision, feng2025video, li2025videochat, wang2025timezero}.

Although GRPO improves training efficiency, it also introduces a critical inefficiency.
Each training instance forms a group of candidates that share a common input prefix (e.g., prompts or instructions). However, during the forward pass, this \textit{shared input prefix} must be re-encoded independently for each group member, resulting in redundant computation that scales with the group size. In long-context reinforcement learning tasks, the prefix often constitutes a substantial portion of the total input sequence. This overhead significantly increases training cost and memory usage, thereby creating a major bottleneck to scalability.

To address this issue, we propose \textbf{\methodname}, an efficient GRPO training algorithm that eliminates redundant prefix computation via a \textit{Shared-Prefix Forward} strategy.
This design enables the shared input prefix to be encoded only once, regardless of the group size, by restructuring attention computation without modifying the model architecture. 
\textbf{Our core idea }is to split the self-attention computation into two parts. One part performs self-attention over only the shared prefix tokens to update their contextual representations. The other computes query embeddings from the suffix tokens, while using the full sequence (prefix + suffix) to compute keys and values. The suffix tokens then attend to both prefix and suffix via a concat-attention mechanism to obtain updated representations. Notably, unlike standard key-value caching mechanisms that are typically designed for inference and do not support gradient backpropagation, our method preserves full differentiability and remains compatible with end-to-end training.

We provide both theoretical and empirical evidence that \methodname is training-equivalent to standard GRPO.
Theoretically, we prove that \methodname yields identical forward outputs and backward gradients to those of the standard GRPO formulation, ensuring that the optimization dynamics and final policy performance remain unchanged. Empirically, we validate this equivalence through extensive experiments, showing that \methodname achieves consistent performance while significantly reducing the computational cost of training. It can be seamlessly integrated into current training pipelines without requiring any structural modifications and only involves modifying a few lines of code related to input construction and attention computation. The computational savings are especially pronounced in scenarios with long shared prefixes.
By reducing redundant computation, \methodname allows for larger group sizes within the same computational budget, thereby enhancing the scalability of GRPO methods to more complex tasks and larger models.

Our main contributions are as follows:
\begin{itemize}
\item We identify a key inefficiency in standard GRPO training: shared input prefixes are repeatedly encoded for each group member, resulting in significant computational overhead that limits scalability in long-context reinforcement learning.
\item We introduce \methodname, a general and implementation-friendly GRPO training algorithm that eliminates redundant prefix computation via a Shared-Prefix Forward strategy. \methodname is fully plug-and-play and compatible with existing GRPO-based architectures.
\item We provide theoretical guarantees of gradient equivalence and computational efficiency. Extensive experiments further demonstrate that \methodname significantly accelerates training in long-prefix scenarios without compromising policy performance.
\end{itemize}

\section{\methodname for GRPO}

As is mentioned above, traditional GRPO implementations suffer from redundant computations when processing long query prefixes (e.g., multi-modal inputs like video tokens), as the same prefix is repeatedly encoded for each group member. This inefficiency becomes pronounced with larger group sizes, which are essential for reducing advantage variance. To address this, we introduce \textbf{\methodname}, a novel algorithm that eliminates redundant prefix computations by leveraging shared-prefix forward. In this section, we will first introduce the algorithm, and then analyze the gradient equivalence and computational savings of \methodname.

\begin{figure*}[t]
\centering
\includegraphics[width=\linewidth]{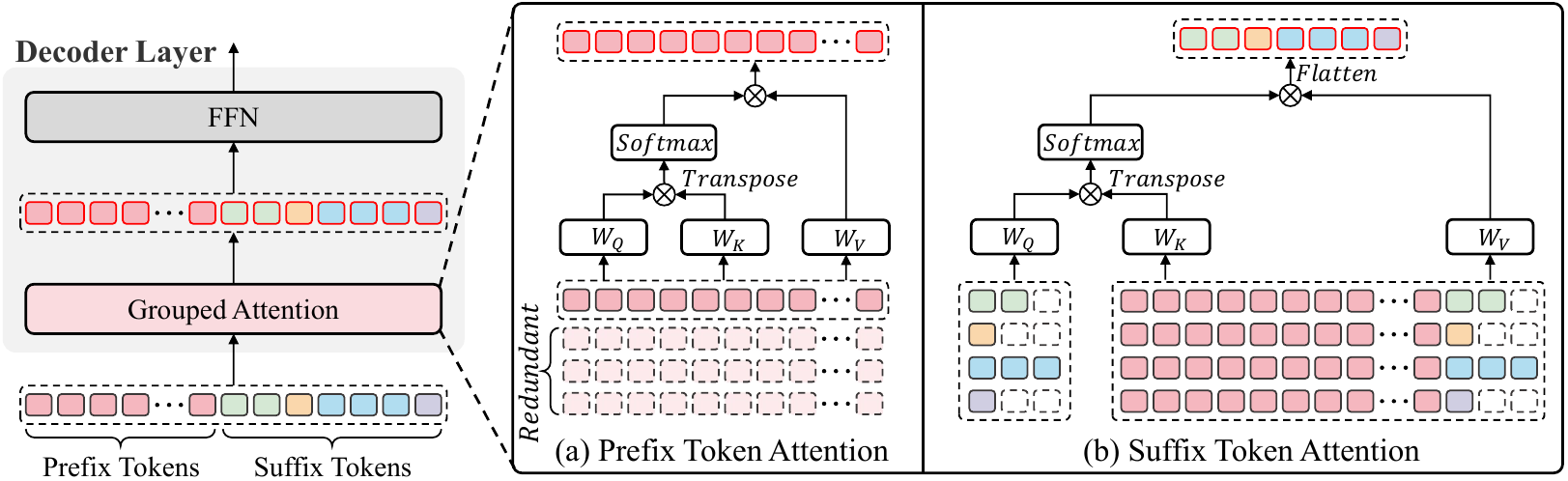}
\caption{Method illustration of Grouped Attention in \methodname.}
\label{fig:method}
\end{figure*}

\subsection{Algorithm implementation}

For clarity, we assume a forward batch size of 1, where the input consists of a single query prefix $P \in \mathbb{R}^{1 \times L \times D}$ (typically containing system prompts, multi-modal inputs, and questions). Given group size $G$, the model samples $G$ response candidates $\{R_1, R_2, \dots, R_{G}\}$ from $P$, where each $R_i \in \mathbb{R}^{1 \times L_i \times D}$. Traditional approaches, which we refer to as Repeated-Prefix Forward, process the inputs as follows:

\begin{equation}
\begin{gathered}
    x_i = [P; R_i], \\
    X_{\text{base}} = \texttt{pad}(x_1, x_2, \dots, x_G),
\end{gathered}
\end{equation}

where \texttt{pad} denotes length-padding operation, resulting in $X_{\text{base}} \in \mathbb{R}^{1 \times (L + \max(L_1, L_2, \dots, L_G)) \times D}$.

In \methodname, we instead concatenate the shared prefix with all response suffixes along the sequence dimension:

\begin{equation}
    X_{\text{ours}} = [P; R_1; R_2; \dots; R_G],
\end{equation}

where $X_{\text{ours}} \in \mathbb{R}^{1 \times (L + L_1 + \dots + L_G) \times D}$. It is obvious that the concatenated samples maintain equivalence with Repeated-Prefix Forward in word embedding and FFN layers.

While Repeated-Prefix Forward directly applies self-attention:

\begin{equation}
    O = \texttt{Attn}(Q, K, V, \texttt{mask}),
\end{equation}

where the $G$ rollouts per prefix lead to redundant computation on the repeated prefix in $X_{\text{base}}$, particularly significant for long prefixes. In contrast, our method decomposes the attention computation into two kernel calls (which we refer to as Grouped Attention):

\begin{equation}
\begin{gathered}
    O_{\text{prefix}} = \texttt{Attn}(Q_{\text{prefix}}, K_{\text{prefix}}, V_{\text{prefix}}, \texttt{mask}_{\text{prefix}}), \\
    O_{\text{suffix}} = \texttt{Attn}(Q_{\text{suffix}}, K_{\text{prefix} + \text{suffix}}, V_{\text{prefix} + \text{suffix}}, \texttt{mask}_{\text{prefix} + \text{suffix}}), \\
    O = \texttt{group}(O_{\text{prefix}}, O_{\text{suffix}}),
\end{gathered}
\end{equation}

where $\texttt{group}(\cdot, \cdot)$ represents concatenating using index select.

For positional encoding (using RoPE or its variants), we set the position id of each token the same as the corresponding position in the Repeated-Prefix Forward. The complete process of Grouped Attention in \methodname is summarized in Algorithm~\ref{alg:prefix_grouper} and Figure~\ref{fig:method}.

\begin{algorithm}[!ht]
\caption{Pseudocode of Grouped Attention in a PyTorch-like style}
\label{alg:prefix_grouper}
\algcomment{
\fontsize{7.2pt}{0em}\selectfont \texttt{attention\_interface}: the attention operation;  \texttt{prefix\_grouper}: implemented using \texttt{torch.autograd.Function}.
}
\definecolor{codeblue}{rgb}{0.25,0.5,0.5}
\lstset{
  backgroundcolor=\color{white},
  basicstyle=\fontsize{7.2pt}{7.2pt}\ttfamily\selectfont,
  columns=fullflexible,
  breaklines=true,
  captionpos=b,
  commentstyle=\fontsize{7.2pt}{7.2pt}\color{codeblue},
  keywordstyle=\fontsize{7.2pt}{7.2pt},
}
\begin{lstlisting}[language=python]
def grouped_attention(self, q, k, v, prefix_grouper, **kwargs):
    """
    q, k, v: Shape [b, num_heads, seq_len, head_dim]. q, k should be pre-processed with RoPE in advance.
    prefix_grouper: A plug-and-play module implemented by us.
    kwargs: Any arguments needed by the attention operation.
    """
    # Split the concatenated samples into prefix and suffix
    q_prefix, k_prefix, v_prefix, q_suffix, k_suffix, v_suffix = prefix_grouper.ungroup(q, k, v)
    # Attention call
    prefix_attn_output, _ = attention_interface(
        self,
        q_prefix,
        k_prefix,
        v_prefix,
        # NOTE: Attention mask is pre-computed by prefix_grouper
        prefix_grouper.prefix_attn_mask.to(q_prefix.device),
        **kwargs,
    )
    suffix_attn_output, _ = attention_interface(
        self,
        q_suffix,
        prefix_grouper.batch_repeat_cat(k_prefix, k_suffix, cat_dim=2),
        prefix_grouper.batch_repeat_cat(v_prefix, v_suffix, cat_dim=2),
        # NOTE: Attention mask is pre-computed by prefix_grouper
        prefix_grouper.suffix_attn_mask.to(q_prefix.device),
        **kwargs,
    )
    # Concatenate the prefix and suffix output
    # The input shape should be [b, seq_len, num_heads, head_dim]
    attn_output = prefix_grouper.group(prefix_attn_output, suffix_attn_output)
    return attn_output, None
\end{lstlisting}
\end{algorithm}

\subsection{Gradient equivalence of \methodname}

A critical property of \methodname is its \textbf{theoretical equivalence} to Repeated-Prefix Forward in both forward outputs and backward gradients. Forward equivalence is self-evident: every token undergoes fundamentally identical computations in both algorithms. Gradient equivalence is demonstrated below:

\begin{coloredlemma}[Gradient Equivalence]
\label{lemma:grad_eq}
Under the Grouped Attention in our \methodname, the gradients of the policy loss function with respect to model parameters $\theta$ are identical to those computed by the original GRPO algorithm, i.e.,
\begin{equation}
    \nabla_\theta \mathcal{J}_{\text{ours}}(X_{\text{ours}}, A) \equiv \nabla_\theta \mathcal{J}_{\text{base}}(X_{\text{base}}, A),
\end{equation}
where $A$ denotes the advantage of the rollout responses.

\textit{Proof.} See Appendix \ref{sec:proof_grad_eq}
\end{coloredlemma}

This property ensures that \methodname's computational efficiency gains come \textbf{without trade-offs}, which means it accelerates training and reduces GPU memory while preserving model performance identically.

\subsection{Computational cost analysis of \methodname}

\methodname employs the Shared-Prefix Forward approach, which demonstrates increasingly significant advantages over Repeated-Prefix Forward as group size grows:

\begin{coloredlemma}[Computation Reduction]
\label{lemma:complexity}
Given a group size $G$ and sequence lengths $L_p$ (prefix), $L_r$ (response), when $L_p \gg L_r$, \methodname reduces FLOPs to $\frac{1}{G}$ of the Repeated-Prefix Forward approach.

\textit{Proof.} See Appendix \ref{sec:proof_complexity}
\end{coloredlemma}

Under long-prefix scenarios (e.g., multi-modal inputs or extended text contexts), \methodname effectively reduces computational load and memory consumption, requiring only marginal computational increase as group size scales, demonstrating its superior efficiency.

\subsection{Futher discussion}

Similar to the GRPO training process, our \methodname can be extended to other shared-prefix inference scenarios requiring computational efficiency. A representative application is multi-QA judge modeling: given a long response $R$, the model processes multiple short questions $\{Q_1, Q_2, \dots, Q_k\}$ sharing $R$ as context. The acceleration is achieved by computing perplexity-based predictions exclusively from hidden states at \textit{final tokens} of each question, enabling single-token prediction for evaluation metrics or rewards.

\section{Experiment}

\subsection{Computational overhead}

\begin{figure}[htbp]
    \centering
    \begin{subfigure}{0.49\textwidth}
        \centering
        \includegraphics[width=\linewidth]{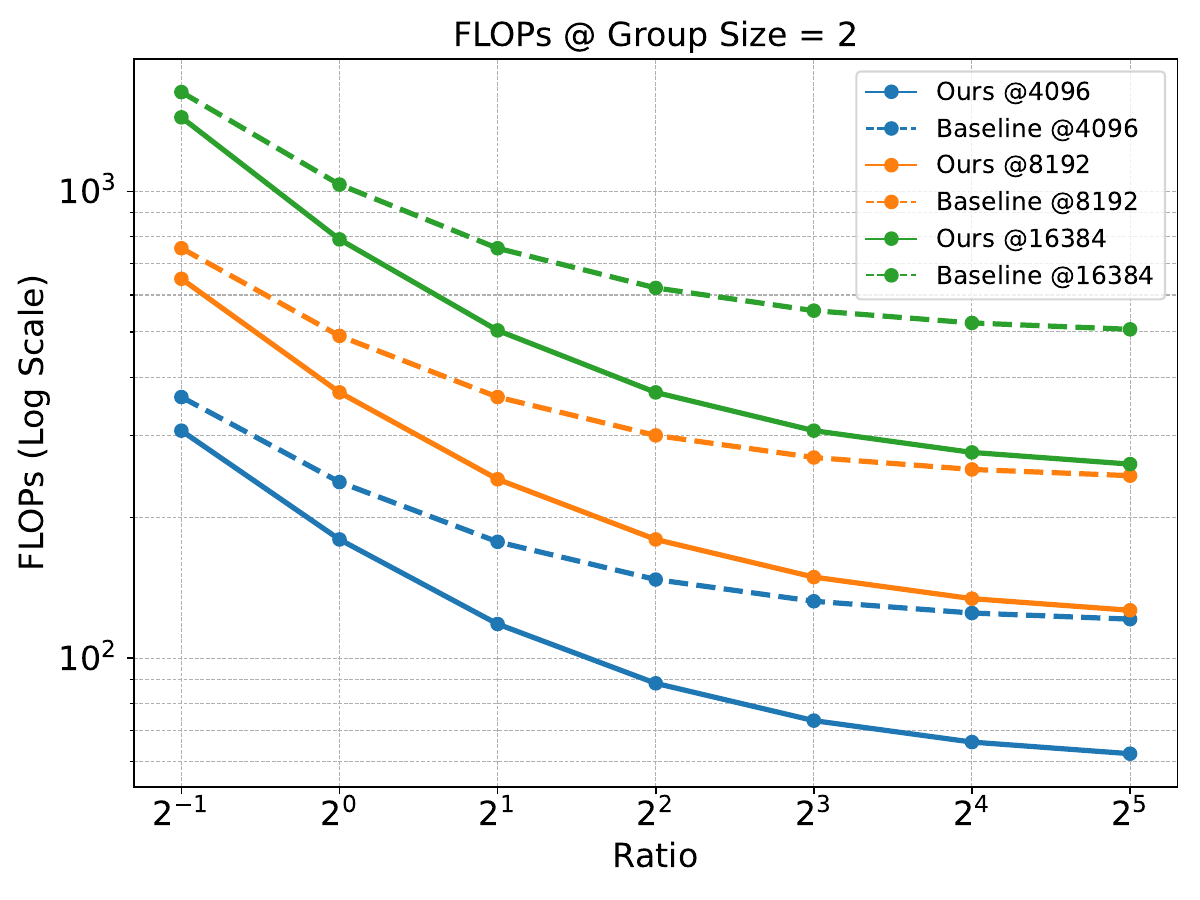}
    \end{subfigure}
    \hfill
    \begin{subfigure}{0.49\textwidth}
        \centering
        \includegraphics[width=\linewidth]{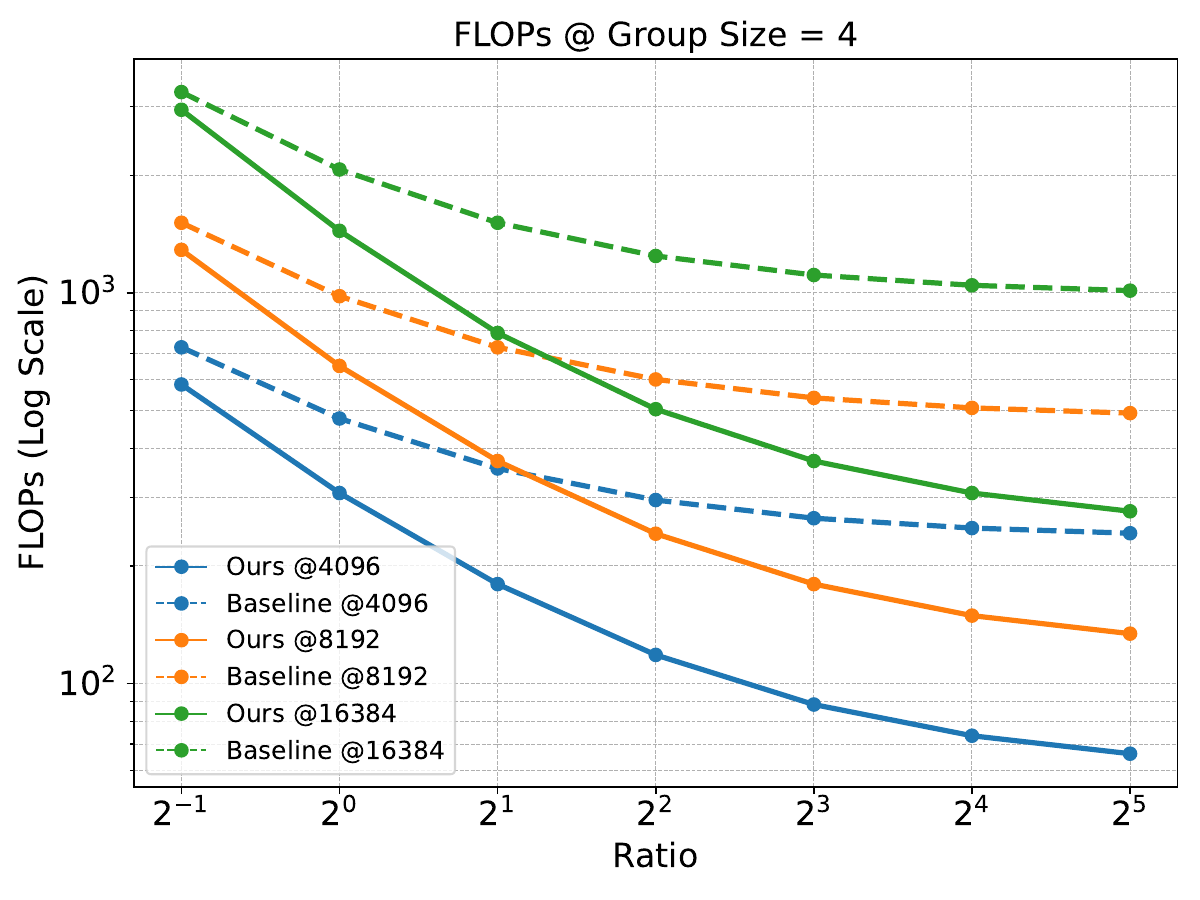}
    \end{subfigure}
    \vspace{0.5cm}
    \begin{subfigure}{0.49\textwidth}
        \centering
        \includegraphics[width=\linewidth]{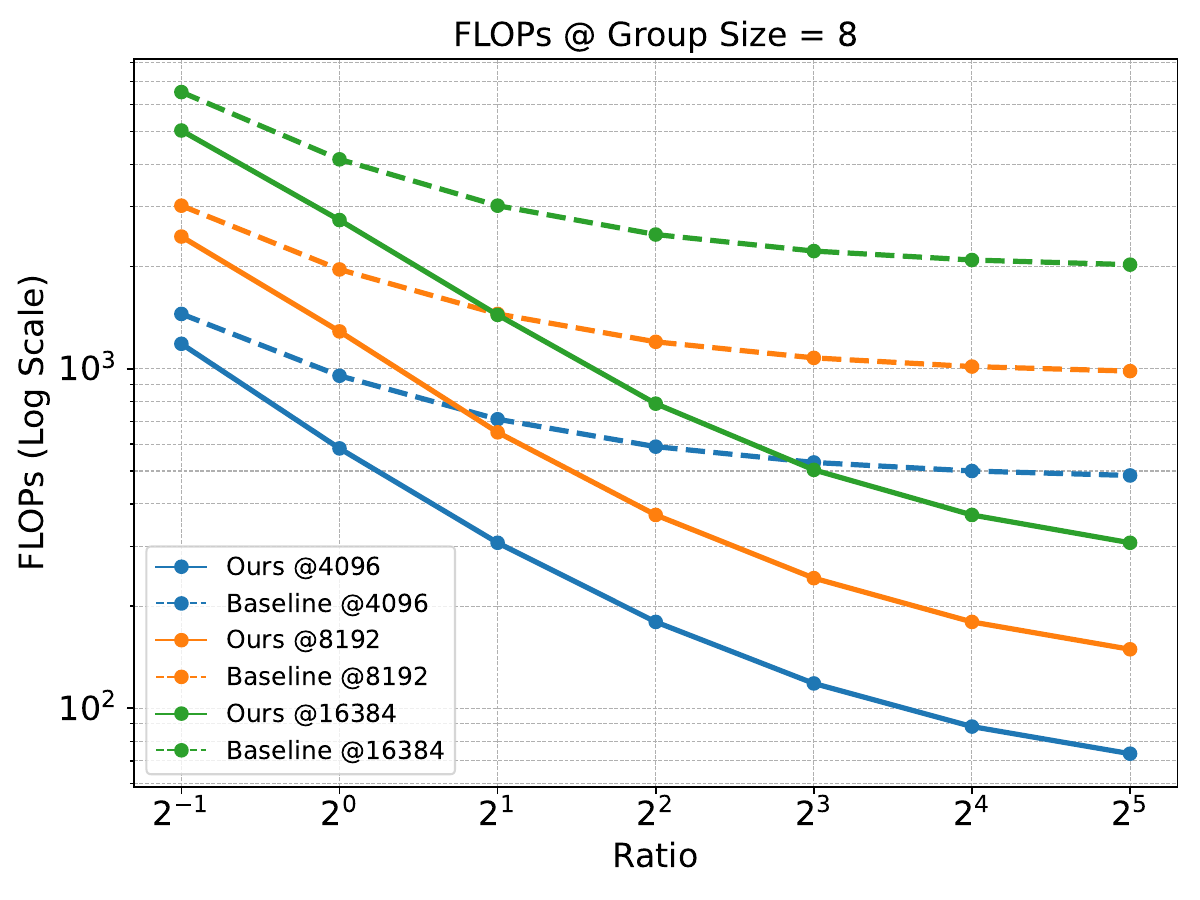}
    \end{subfigure}
    \hfill
    \begin{subfigure}{0.49\textwidth}
        \centering
        \includegraphics[width=\linewidth]{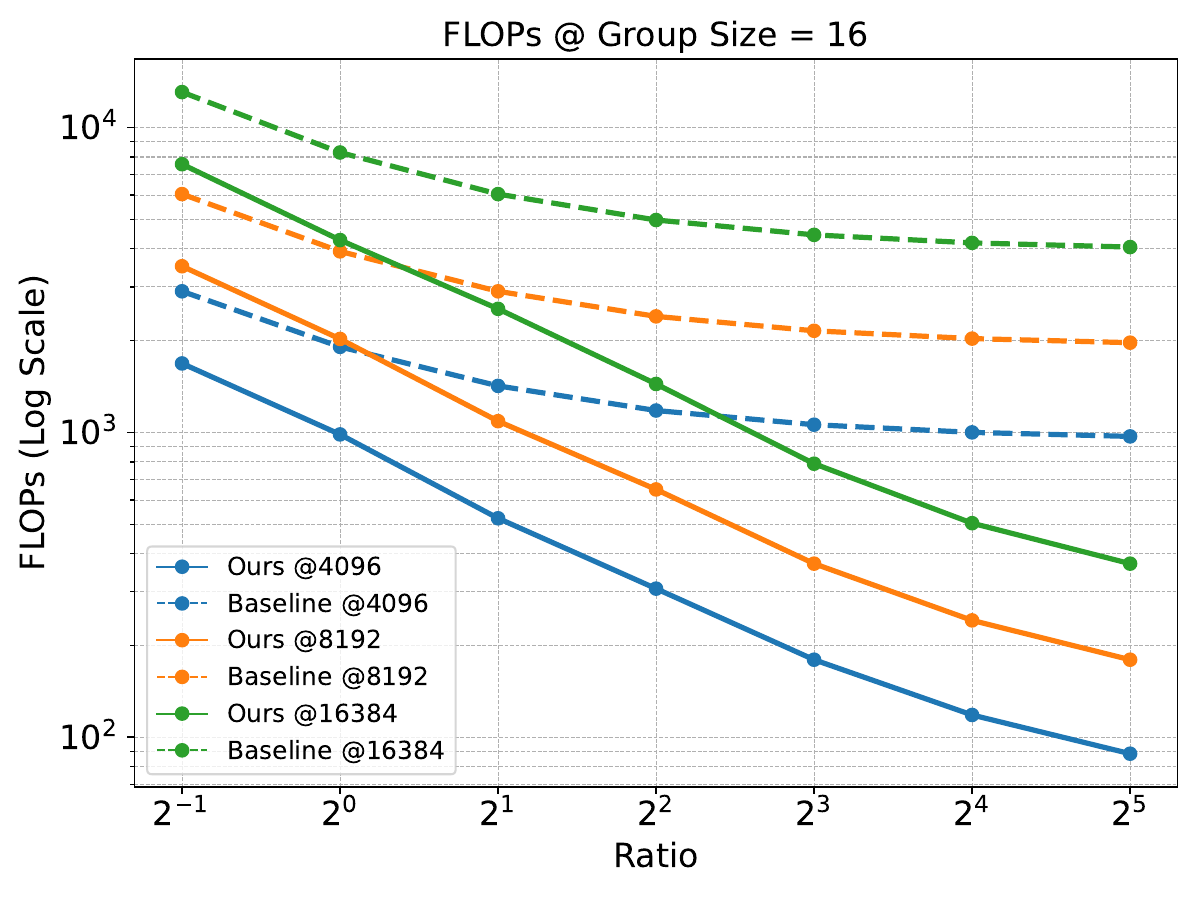}
    \end{subfigure}
    \caption{Comparison of FLOPs under different group sizes. The figure displays results at fixed prefix lengths (4096, 8192, and 16384) across different Ratios (prefix length / suffix length).}
    \label{fig:flops}
\end{figure}

Comparison of computational overhead is shown in Figure \ref{fig:flops}. Within any given group size investigated (2, 4, 8, or 16), \methodname ("Ours") consistently demonstrates a significant computational advantage over the "Baseline" approach. This superiority is evidenced by the markedly lower Floating Point Operations (FLOPs) required by our method across all tested configurations (@4096, @8192, @16384) and ratios. The consistent reduction in computational load for the same group size underscores the enhanced efficiency and practical benefits of our proposed technique.

\subsection{Memory usage}

\begin{figure}[htbp]
    \centering
    \begin{subfigure}{0.49\textwidth}
        \centering
        \includegraphics[width=\linewidth]{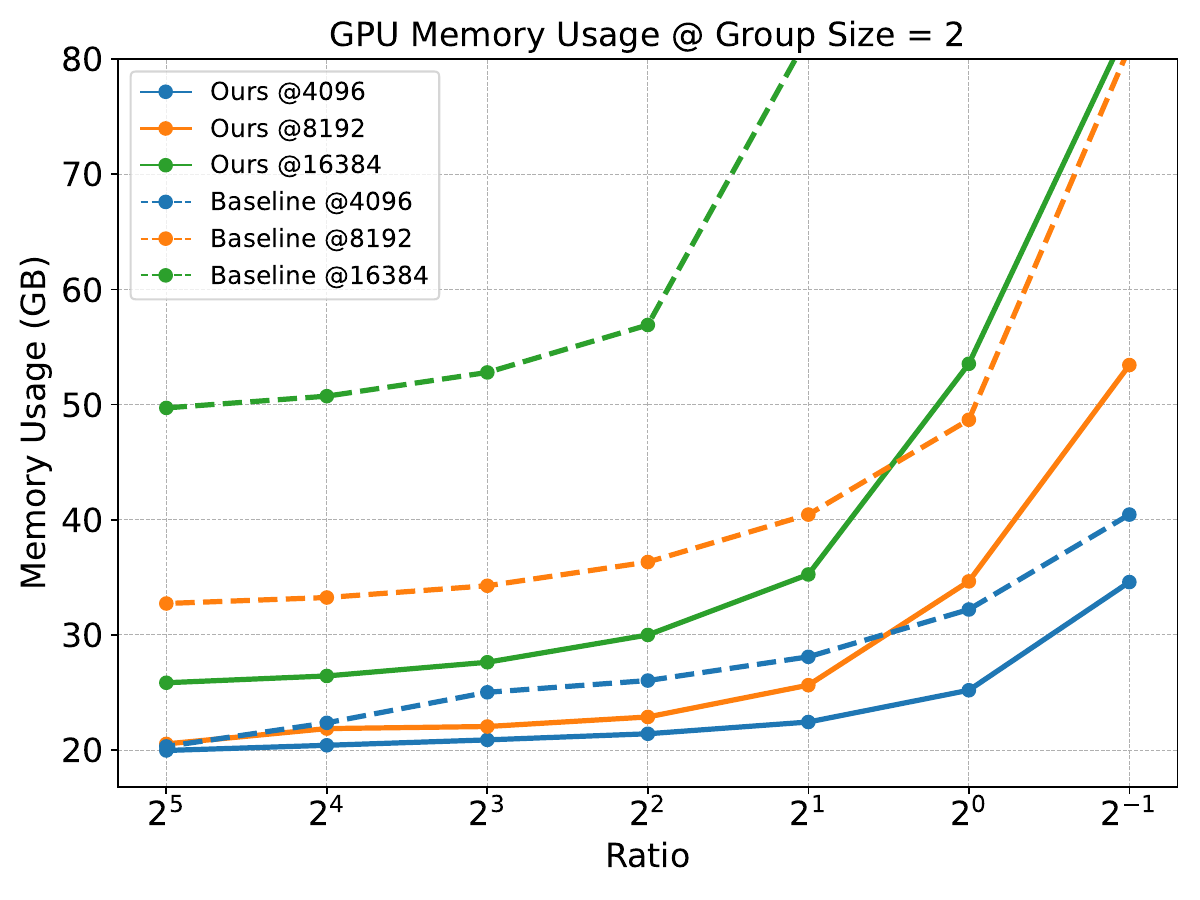}
    \end{subfigure}
    \hfill
    \begin{subfigure}{0.49\textwidth}
        \centering
        \includegraphics[width=\linewidth]{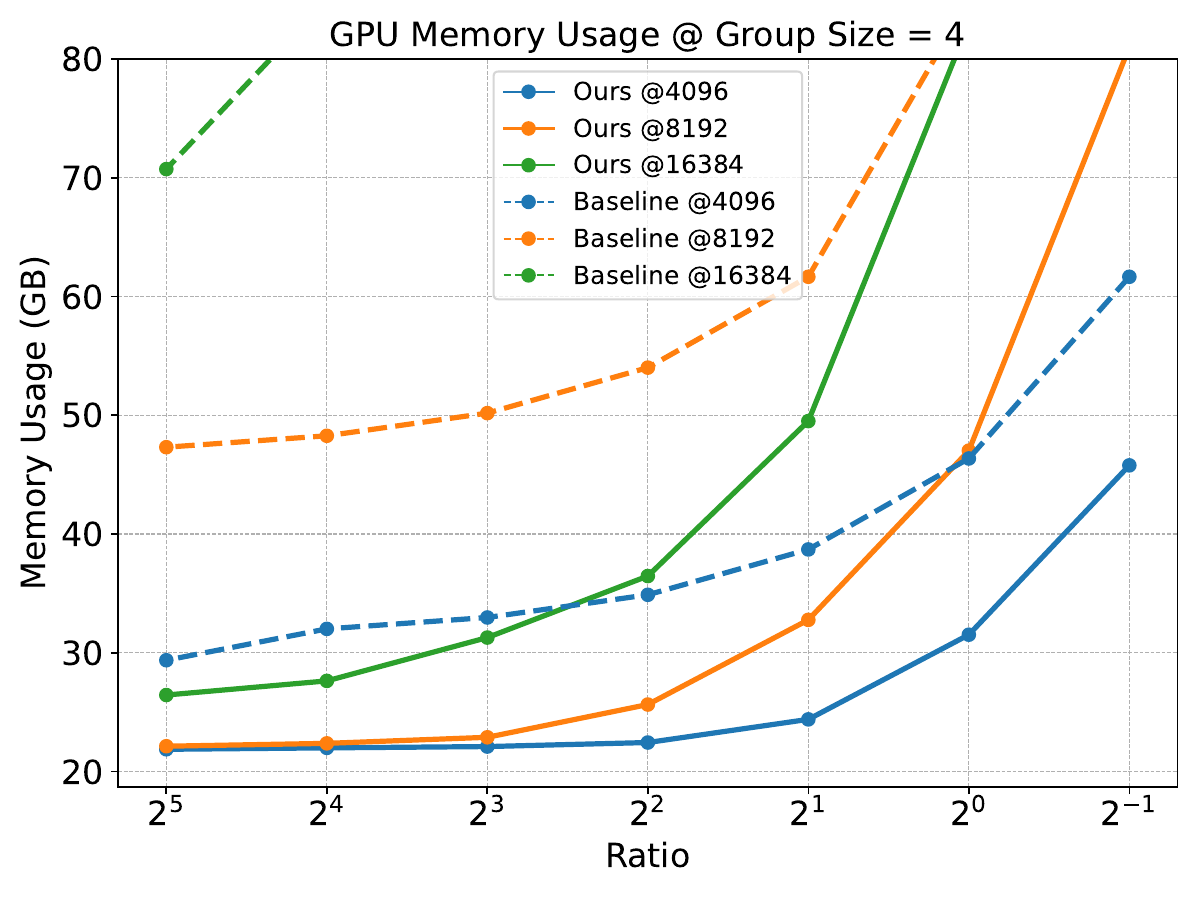}
    \end{subfigure}
    \vspace{0.5cm}
    \begin{subfigure}{0.49\textwidth}
        \centering
        \includegraphics[width=\linewidth]{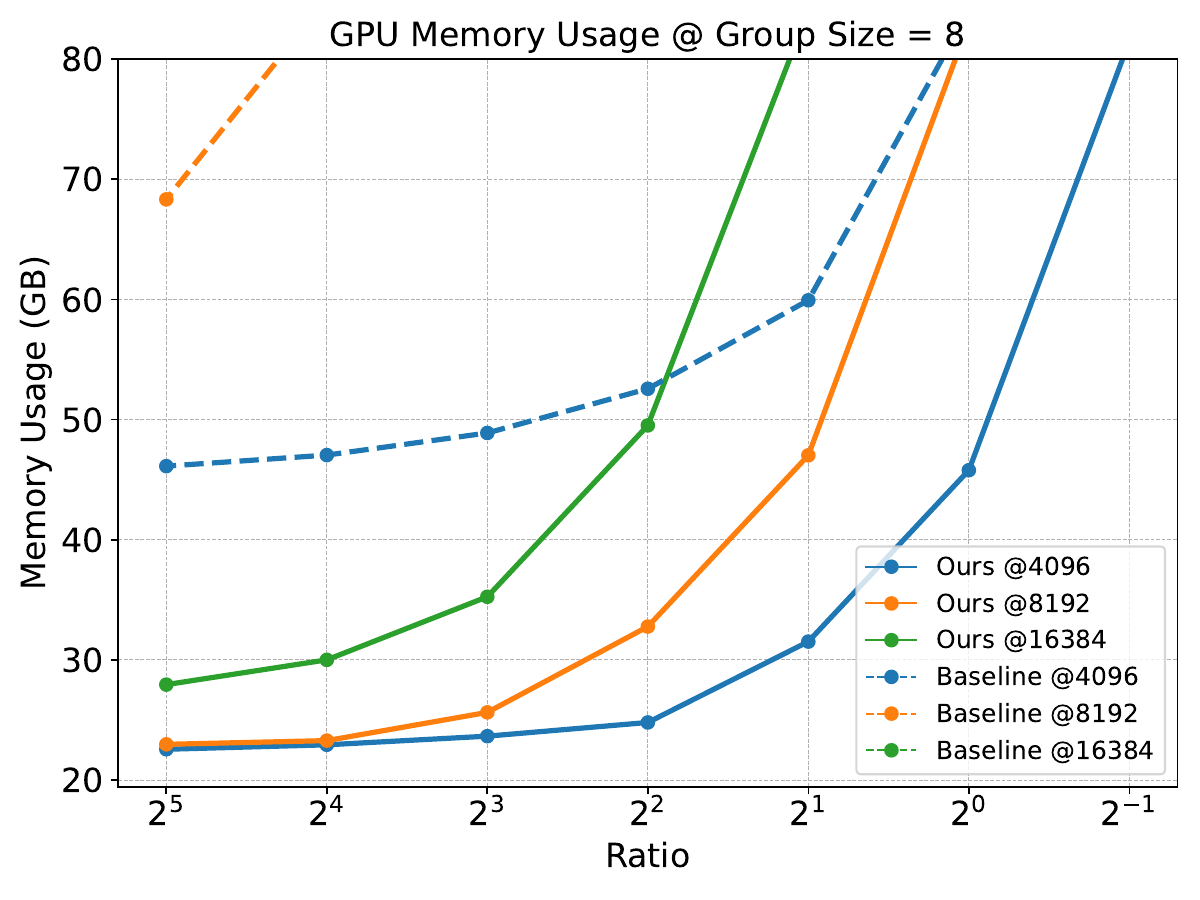}
    \end{subfigure}
    \hfill
    \begin{subfigure}{0.49\textwidth}
        \centering
        \includegraphics[width=\linewidth]{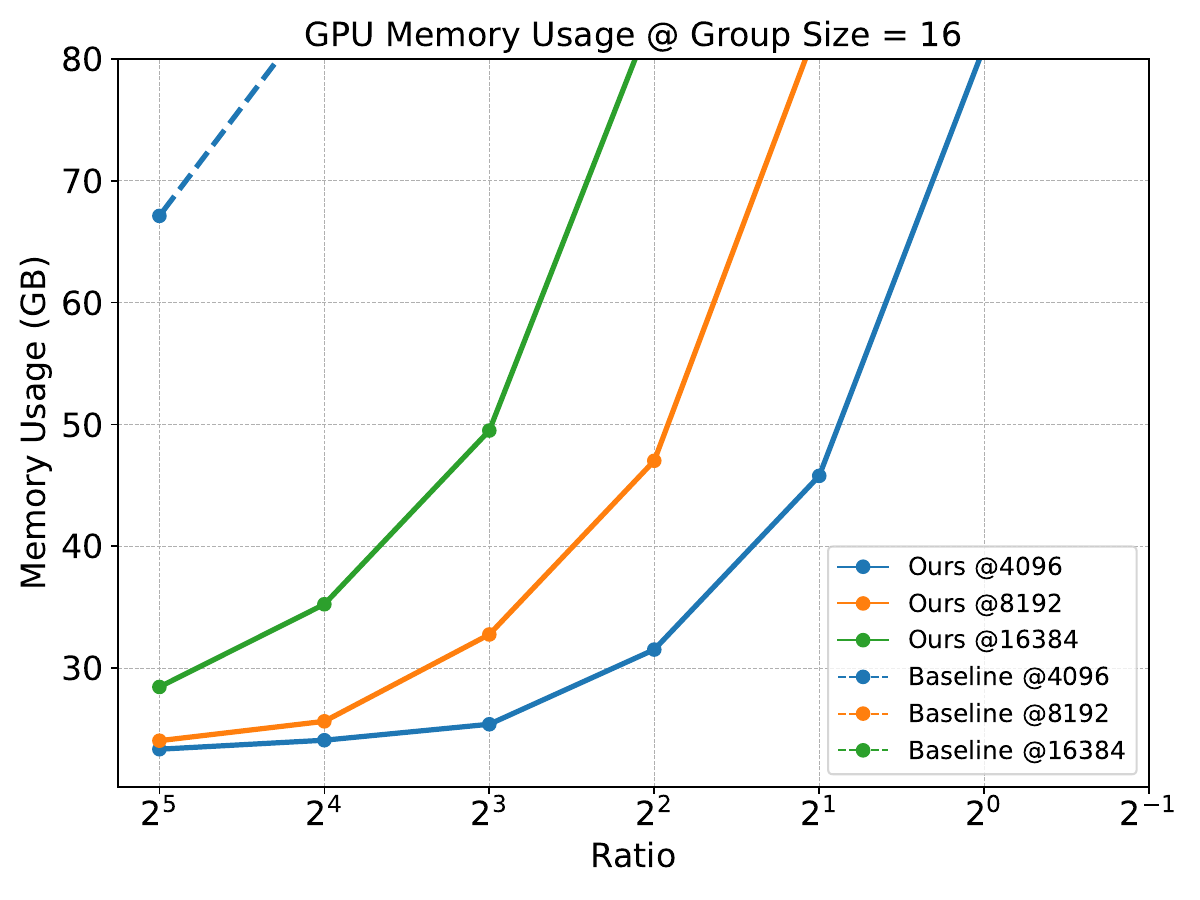}
    \end{subfigure}
    \caption{Comparison of memeory usage under different group sizes. The figure displays results at fixed prefix lengths (4096, 8192, and 16384) across different Ratios (prefix length / suffix length).}
    \label{fig:mem}
\end{figure}

Figure \ref{fig:mem} illustrates a comparative analysis of GPU memory usage between \methodname ("Ours") and the "Baseline" approach, evaluated under identical group size conditions of 2, 4, 8, and 16. Across all tested group sizes, a clear and consistent pattern indicates that our method achieves a substantial reduction in GPU memory consumption when compared directly to the baseline for all configurations (@4096, @8192, @16384) and over the entire range of ratios. Our method consistently maintains a significantly lower memory footprint.

\section{Conclusion}

We present \methodname, an efficient algorithm for Group Relative Policy Optimization (GRPO) that eliminates redundant computation of shared input prefixes through a novel Shared-Prefix Forward strategy. By restructuring attention into two kernel calls—encoding prefixes once while enabling suffixes to attend to full contexts—our method achieves identical forward outputs and backward gradients to standard GRPO, preserving optimization dynamics. This implementation-agnostic approach substantially reduces training costs, particularly for long-context tasks, and serves as a drop-in replacement to enhance GRPO’s scalability without compromising policy performance.

{
\small
\bibliographystyle{unsrt}
\bibliography{reference}
}

\newpage

\appendix

\section{Proof}

\subsection{Proof of Gradient Equivalence}
\label{sec:proof_grad_eq}

\begin{coloredproof}[(1 of 2)]
\label{proof:grad_eq}
For clarity, we assume batch size $B=1$. For a transformer with parameters $\theta$, the total gradient $\nabla_\theta \mathcal{J}$ is the sum of per-token contributions:
\begin{equation}
    \nabla_\theta \mathcal{J} = \sum_{t=1}^T \underbrace{\frac{\partial \mathcal{J}}{\partial \mathbf{h}_t} \cdot \nabla_\theta \mathbf{h}_t}_{\text{token } t \text{'s gradient contribution}},
\end{equation}
\noindent where $\mathbf{h}_t$ is the hidden state at position $t$. This holds under additive loss decomposition: $\mathcal{J} = \sum_t \mathcal{J}_t(\mathbf{h}_t)$ (e.g., next-token prediction)

We decompose the transformer architecture into two computational categories: 1. \textbf{Attention operations}: $O = \mathrm{Attn}(Q, K, V)$. 2. \textbf{Pointwise operations}: MLP, QKV projections, and normalization layers.

All learnable parameters reside in pointwise operations. Gradient equivalence is established by analyzing each category separately.

\subsection*{Part 1: Attention Gradient}
Consider the attention output $O$ and its gradients $\nabla_{Q,K,V} O$. For suffix tokens ($R_i$), both implementations compute identical gradients due to equivalent computational paths:
\begin{equation}
    \nabla_{Q_{\mathrm{suffix}}, K_{\mathrm{suffix}}, V_{\mathrm{suffix}}} O_{\text{ours}} \equiv \nabla_{Q_{\mathrm{suffix}}, K_{\mathrm{suffix}}, V_{\mathrm{suffix}}} O_{\text{base}}.
\end{equation}

For prefix tokens ($P$), the gradient is calculated as follows:

\begin{align}
    \nabla_{P_i(Q, K, V)} O_{\text{base}} &= \underbrace{\nabla_{P_i(Q, K, V)} O_i^{\text{prefix}}}_{\text{prefix-only}} + \underbrace{\nabla_{P_i(Q, K, V)} O_i^{\text{suffix}}}_{\text{response interaction}}, \\
    \nabla_{P(Q, K, V)} O_{\text{ours}} &= \nabla_{P(Q, K, V)} O^{\text{prefix}} + \frac{1}{G} \sum_{i=1}^G \left(\nabla_{P(Q, K, V)} O_i^{\text{suffix}} \right).
\end{align}

\subsection*{Part 2: Pointwise Operation Gradient}
For any pointwise operation parameter $\theta$, gradients are computed as:
\begin{align}
    \nabla_\theta \mathcal{J}_{\text{base}} &= \frac{1}{G} \sum_{i=1}^G \left( \sum_{t \in P_i} \frac{\partial \mathcal{J}_{\text{base}}}{\partial \mathbf{h}_t} \nabla_\theta \mathbf{h}_t + \sum_{t \in R_i} \frac{\partial \mathcal{J}_{\text{base}}}{\partial \mathbf{h}_t} \nabla_\theta \mathbf{h}_t \right), \label{eq:base_grad} \\
    \nabla_\theta \mathcal{J}_{\text{ours}} &= \sum_{t \in P} \frac{\partial \mathcal{J}_{\text{ours}}} {\partial \mathbf{h}_t} \nabla_\theta \mathbf{h}_t + \frac{1}{G} \sum_{i=1}^G \sum_{t \in R_i} \frac{\partial \mathcal{J}_{\text{ours}}}{\partial \mathbf{h}_t} \nabla_\theta \mathbf{h}_t. \label{eq:ours_grad}
\end{align}

In the final output, the GRPO loss $\mathcal{J}$ depends \textit{only} on response tokens $R_i$:
\begin{equation}
    \forall t \in P,\ \frac{\partial \mathcal{J}}{\partial \mathbf{h}_t} = 0 \quad \text{(both algorithms)},
\end{equation}
\noindent so based on Eq. \ref{eq:base_grad} and Eq. \ref{eq:ours_grad}, the gradients of the final FFN and output embedding layers are equivalent in both algorithms.

For the attention part, we have:

\begin{equation}
\label{eq:attn_eq}
    \frac{1}{G} \sum_{i=1}^{G} \nabla_{P_i(Q, K, V)} O_{\text{base}} = \nabla_{P(Q, K, V)} O_{\text{ours}},
\end{equation}
\end{coloredproof}

\begin{coloredproof}[(2 of 2)]
\label{proof:grad_eq2}
so substitute Eq. \ref{eq:attn_eq} into Eq. \ref{eq:base_grad} and Eq. \ref{eq:ours_grad}, we have:
\begin{align}
\nabla_\theta \mathcal{J}_{\text{base}} &= \frac{1}{G} \sum_{i=1}^G \left( \sum_{t \in P_i} \frac{\partial \mathcal{J}_{\text{base}}}{\partial \mathbf{h}_t} \nabla_\theta \mathbf{h}_t + \sum_{t \in R_i} \frac{\partial \mathcal{J}_{\text{base}}}{\partial \mathbf{h}_t} \nabla_\theta \mathbf{h}_t \right) \nonumber\\
&= \frac{1}{G} \sum_{i=1}^G \sum_{t \in P_i} \frac{\partial \mathcal{J}_{\text{base}}}{\partial \mathbf{h}_t} \nabla_\theta \mathbf{h}_t + \frac{1}{G} \sum_{i=1}^G \sum_{t \in R_i} \frac{\partial \mathcal{J}_{\text{base}}}{\partial \mathbf{h}_t} \nabla_\theta \mathbf{h}_t \nonumber\\
&= \sum_{t \in P} \frac{\partial \mathcal{J}_{\text{ours}}} {\partial \mathbf{h}_t} \nabla_\theta \mathbf{h}_t + \frac{1}{G} \sum_{i=1}^G \sum_{t \in R_i} \frac{\partial \mathcal{J}_{\text{ours}}}{\partial \mathbf{h}_t} \nabla_\theta \mathbf{h}_t \nonumber\\
&= \nabla_\theta \mathcal{J}_{\text{ours}}.
\end{align}

Therefore, through layer-by-layer backpropagation, gradient equivalence propagates upstream from output layers. Thus, $\nabla_\theta \mathcal{J}_{\text{ours}}(X_{\text{ours}}, A) \equiv \nabla_\theta \mathcal{J}_{\text{base}}(X_{\text{base}}, A)$ for all parameters $\theta$.\qedhere
\end{coloredproof}

\subsection{Proof of Computation Reduction}
\label{sec:proof_complexity}

\begin{coloredproof}
\label{proof:complexity}
For clarity, we assume batch size $B=1$ and uniform response length $L_r$ across all $G$ responses. Let $L_p$ denote prefix length, $d$ head dimension, and $n$ number of attention heads. The computational complexity is analyzed separately for causal attention and pointwise operations (MLP \& QKV projections).

\noindent\textbf{Causal Attention Operation:}
The baseline Repeated-Prefix Forward method computes:
\begin{equation}
\label{eq:attn_baseline}
\mathcal{C}_{\text{attn}}^{\text{base}} = G(L_p + L_r)^2 d n
\end{equation}
Our \methodname decomposes attention into prefix and suffix components:
\begin{equation}
\label{eq:attn_ours}
\mathcal{C}_{\text{attn}}^{\text{ours}} = \underbrace{L_p^2 d n}_{\text{prefix self-attn}} + \underbrace{G L_r (2L_p + L_r) d n}_{\text{suffix attn}}
\end{equation}
The complexity ratio simplifies to:
\begin{align}
\frac{\mathcal{C}_{\text{attn}}^{\text{ours}}}{\mathcal{C}_{\text{attn}}^{\text{base}}} 
&= \frac{L_p^2 d n + G L_r (2L_p + L_r) d n}{G(L_p + L_r)^2 d n} \nonumber \\
&= \frac{L_p^2 + G L_r (2L_p + L_r)}{G(L_p + L_r)^2}
\end{align}
As $L_p \gg L_r$, the asymptotic limit is:
\begin{equation}
\lim_{L_p/L_r \to \infty} \frac{\mathcal{C}_{\text{attn}}^{\text{ours}}}{\mathcal{C}_{\text{attn}}^{\text{base}}} = \frac{L_p^2}{G L_p^2} = \frac{1}{G}
\end{equation}

\noindent\textbf{Pointwise Operation:}
Let $\mathcal{C}_{\text{ffn}}$ denote FLOPs per token for MLP and projections. The baseline requires:
\begin{equation}
\mathcal{C}_{\text{pointwise}}^{\text{base}} = G(L_p + L_r) \mathcal{C}_{\text{ffn}}
\end{equation}
while \methodname computes:
\begin{equation}
\mathcal{C}_{\text{pointwise}}^{\text{ours}} = L_p \mathcal{C}_{\text{ffn}} + G L_r \mathcal{C}_{\text{ffn}}
\end{equation}
The asymptotic ratio is:
\begin{equation}
\lim_{L_p/L_r \to \infty} \frac{\mathcal{C}_{\text{pointwise}}^{\text{ours}}}{\mathcal{C}_{\text{pointwise}}^{\text{base}}} = \lim_{L_p/L_r \to \infty} \frac{L_p + G L_r}{G(L_p + L_r)} = \frac{1}{G}
\end{equation}

\noindent\textbf{Conclusion:} Both attention and pointwise operations exhibit $\mathcal{O}(1/G)$ complexity reduction under $L_p \gg L_r$ conditions.
\end{coloredproof}
\end{document}